\documentclass[letterpaper, 10 pt, conference]{ieeeconf}  

\IEEEoverridecommandlockouts                              

\usepackage{epsfig}
\usepackage{amsmath}
\usepackage{cleveref}
\usepackage{footmisc}
\usepackage{color}
\usepackage{multirow}
\usepackage{textcomp}
\usepackage{amssymb}
\usepackage{array}
\usepackage{stfloats}

\title{ACNN: a Full Resolution DCNN for Medical Image Segmentation}

\author{Xiao-Yun Zhou$^{1*}$ and Jian-Qing Zheng$^{1*}$ and Peichao Li$^{1}$ and Guang-Zhong Yang$^{1,2}$
\thanks{*Xiao-Yun Zhou and Jian-Qing Zheng contribute equally to this paper}
\thanks{$^{1}$The Hamlyn Centre for Robotic Surgery, Imperial College London, UK.
        {\tt\small xiaoyun.zhou14@imperial.ac.uk}}%
\thanks{$^{2}$Institute of Medical Robotics, Shanghai Jiao Tong University, China}%
}

\begin{document}

\maketitle
\thispagestyle{empty}
\pagestyle{empty}

\begin{abstract}
Deep Convolutional Neural Networks (DCNNs) are used extensively in medical image segmentation and hence 3D navigation for robot-assisted Minimally Invasive Surgeries (MISs). However, current DCNNs usually use down sampling layers for increasing the receptive field and gaining abstract semantic information. These down sampling layers decrease the spatial dimension of feature maps, which can be detrimental to image segmentation. Atrous convolution is an alternative for the down sampling layer. It increases the receptive field whilst maintains the spatial dimension of feature maps. In this paper, a method for effective atrous rate setting is proposed to achieve the largest and fully-covered receptive field with a minimum number of atrous convolutional layers. Furthermore, a new and full resolution DCNN - Atrous Convolutional Neural Network (ACNN), which incorporates cascaded atrous II-blocks, residual learning and Instance Normalization (IN) is proposed. Application results of the proposed ACNN to Magnetic Resonance Imaging (MRI) and Computed Tomography (CT) image segmentation demonstrate that the proposed ACNN can achieve higher segmentation Intersection over Unions (IoUs) than U-Net and Deeplabv3+, but with reduced trainable parameters.
\end{abstract}

\section{Introduction}
Medical image segmentation which predicts the class, anatomy, or prosthesis of each pixel in an image is important for robot-assisted Minimally Invasive Surgeries (MISs). For example, the segmentation of Right Ventricle (RV) is important for instantiating the intro-operative 3D RV shape for navigating robot-assisted Radio-frequency Cardiac Ablation (RFCA) in 3D \cite{zhou2018real, zhou2019one}. The segmentation of markers in \cite{zhou2018towards} is essential for instantiating the intra-operative 3D stent graft shape at fully-compressed \cite{zhoustent}, partially-deployed \cite{zheng2019real} and fully-deployed \cite{zhou2018realRAL} state for 3D navigation in Fenestrated Endovascular Aortic Repair (FEVAR). 

\begin{figure}[thpb]
    \centering
    \framebox{\includegraphics[width=0.39\textwidth]{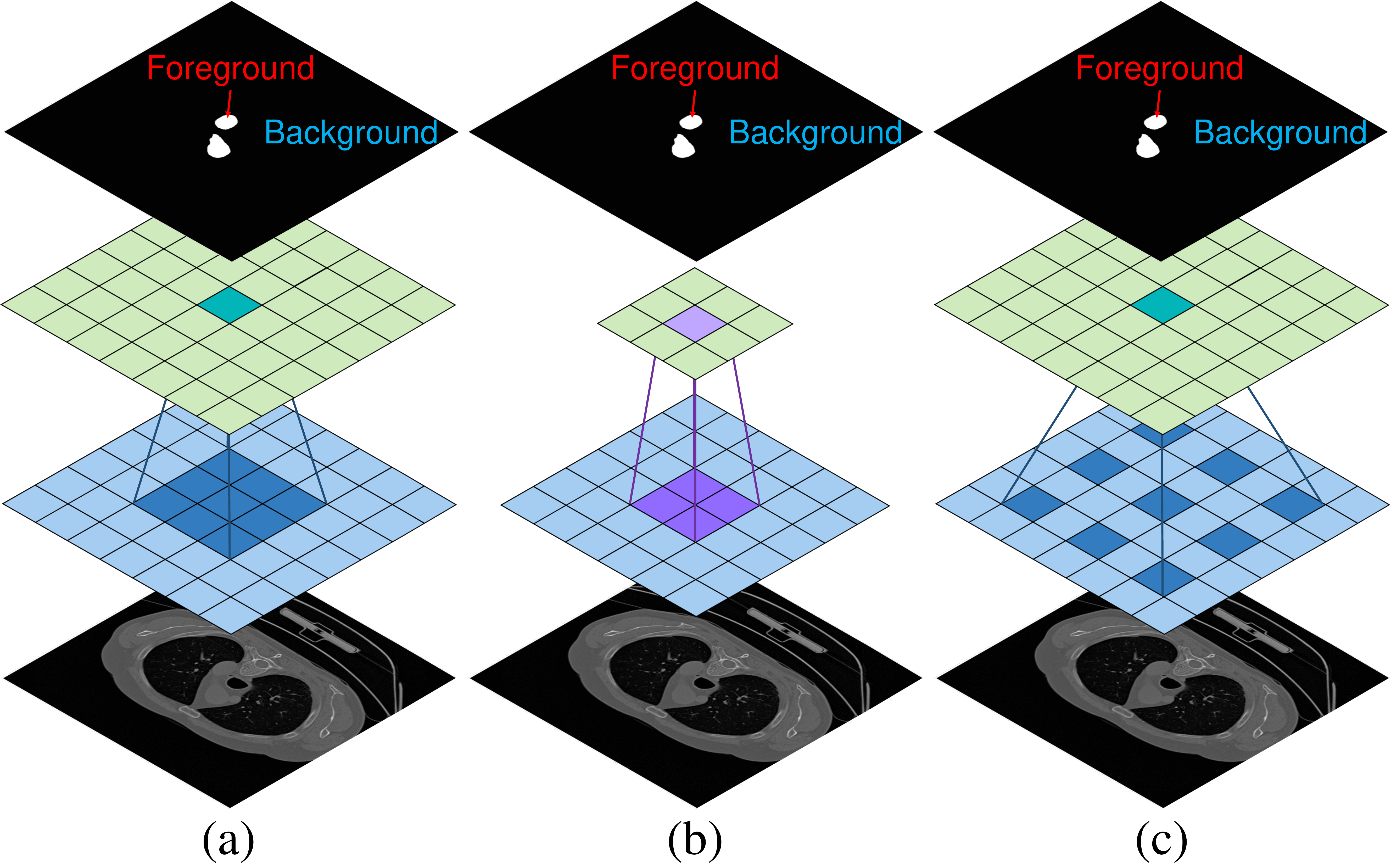}}
    \caption{Illustrations of using DCNN with different receptive fields for medical image segmentation: (a) convolutional layer with a $3\times 3$ receptive field; (b) pooling layer with a $2\times 2$ receptive field; (c) atrous convolutional layer (atrous rate is 2) with a $5\times 5$ receptive field.}
    \label{fig: Intro}
\end{figure}

Conventional methods are based on ad hoc, expert-designed feature extractors and classifiers. Recently, the use of Deep Convolutional Neural Networks (DCNNs) has shown promising results for many vision-based tasks including image classification \cite{krizhevsky2012imagenet}, object detection \cite{lin2017focal}, and semantic image segmentation \cite{chen2018deeplab}. In DCNN, features are extracted and classified automatically by training multiple non-linear modules \cite{lecun2015deep}. Unlike traditional fully-connected neural networks where each output node is linked to all input nodes, an output node of DCNN only links to regional input nodes, known as the receptive field (the input nodes that an output node sees). Multiple convolutional layers, as shown in Fig. \ref{fig: Intro}a, and down sampling layers, i.e., pooling layers shown in Fig. \ref{fig: Intro}b, are cascaded to achieve a large receptive field coverage. The use of this kind of DCNN means that the feature map is also down sampled, which can be detrimental to pixel-level tasks, i.e., image segmentation. For medical images with focal lesions, local features with small sizes may be discarded due to the down sampling.

In order to compensate for decreased dimension of feature maps, various techniques have been proposed. For example, deconvolutional layers and non-linear up-sampling are used respectively in Fully Convolutional Neural Network (FCNN) \cite{long2015fully} and SegNet \cite{badrinarayanan2017segnet} to recover the down-sampled feature map to the input image size. An alternative is to use atrous convolution \cite{chen2018deeplab}, also known as dilated convolution \cite{Yu2016Multiscale}, to replace the down sampling layer in traditional DCNNs to increase the receptive field. Atrous convolution inserts zeros between non-zero filter taps to sample the feature map as shown in Fig. \ref{fig: Intro}c. It increases the receptive field with the atrous rate but maintains the spatial dimension of feature maps without increasing the computational complexity. However, applying atrous convolution introduces a high demand on memory usage and the inserted zeros of atrous convolution cause input node or information missing. These challenges have limited the practical usage of atrous convolution, particularly for medical image segmentation.

As mentioned above, memory shortage is the first challenge for applying atrous convolution, as high-resolution feature map propagation consumes a large amount of memory. In previous work, atrous convolution was usually applied jointly with down sampling layers as a trade-off between the accuracy and memory. For example, in Deeplab \cite{chen2018deeplab}, atrous convolutional layers were applied on a feature map down-sampled at $\frac{1}{8}$ spatial size of the input image. In multi-scale context aggregation \cite{Yu2016Multiscale}, a feature map with $64\times 64$ dimension was firstly down-sampled from the input image, then a context module with seven atrous convolutional layers was applied. Similar joint usage of atrous convolutional and down sampling layers can also be found in \cite{wang2018understanding}.

In practice, setting the atrous rates is another challenge when applying atrous convolution. This is because the output node only links to input nodes which align with non-zero filter taps, as shown in Fig. \ref{fig: Intro}c. The input nodes which align with zero filter taps are not considered. There are thus far no standard ways of setting the atrous rates. For example, an atrous rate setting of (1, 1, 2, 4, 8, 16, 1), representing the atrous rates of seven layers respectively, was allocated for achieving a receptive field of $67\times 67$ in \cite{Yu2016Multiscale} following the strides of max-pooling layers in FCNN. \textit{Wang et al.} found that an atrous rate setting of (2, 4, 8) would cause gridding effects (regular input nodes are missed) and proposed a hybrid atrous rate setting, i.e., (1, 2, 5, 9) to guarantee a coverage of all input nodes \cite{wang2018understanding}. An atrous rate setting of (6, 12, 18) was used for the block and an atrous rate setting of (1, 2, 4) was set inside each block based on empirical knowledge \cite{chen2017rethinking}.

In this paper, we propose a full resolution DCNN where the spatial dimension of intermediate feature maps remains the same as that of the input image. This is different from the work of \cite{pohlen2017full}, for which the spatial dimension of intermediate feature maps at the residual stream is still smaller than that of the input image. For proposing a full resolution DCNN, the proposed network needs to: 1) maximize the receptive field with as few atrous convolutional layers as possible to save memory usage; 2) fully cover the receptive field without missing any input node. In Sec. \ref{sec: Method}, we first prove a method that sets the atrous rate as $(k)^{n-1}$ at the $n^{\rm th}$ atrous convolutional layer, where $k$ is the kernel size and $n$ is the sequence number of atrous convolutional layer, can achieve the largest and fully-covered receptive field with a minimum number of atrous convolutional layers. Due to the truncation effect and for containing more trainable parameters, a full resolution DCNN - Atrous Convolutional Neural Network (ACNN) is proposed by using multiple cascaded atrous II-blocks, residual learning and Instance Normalization (IN). Cardiovascular Magnetic Resonance Imaging (MRI) and Computed Tomography (CT) image segmentation of the RV, Left Ventricle (LV) and aorta are used to validate the proposed ACNN with data collection shown in Sec. \ref{sec: Data} and with results shown in Sec. \ref{sec: Result}. U-Net \cite{ronneberger2015u} and Deeplab \cite{chen2018encoder} are used as the comparison methods for performance assessment. It has been shown that the proposed ACNN can achieve higher segmentation Intersection over Union (IoU) compared to other techniques with much less trainable parameters and model sizes, indicating the benefit of full resolution feature maps in DCNN. Discussions and conclusions are stated in Sec. \ref{sec: Discussion} and Sec. \ref{sec: Conclusion} respectively.

\section{Methodology}
\label{sec: Method}

\subsection{Atrous Rate Setting}
\label{sec: Method-rate}
\begin{figure}[thpb]
    \centering
    \framebox{\includegraphics[width=0.45\textwidth]{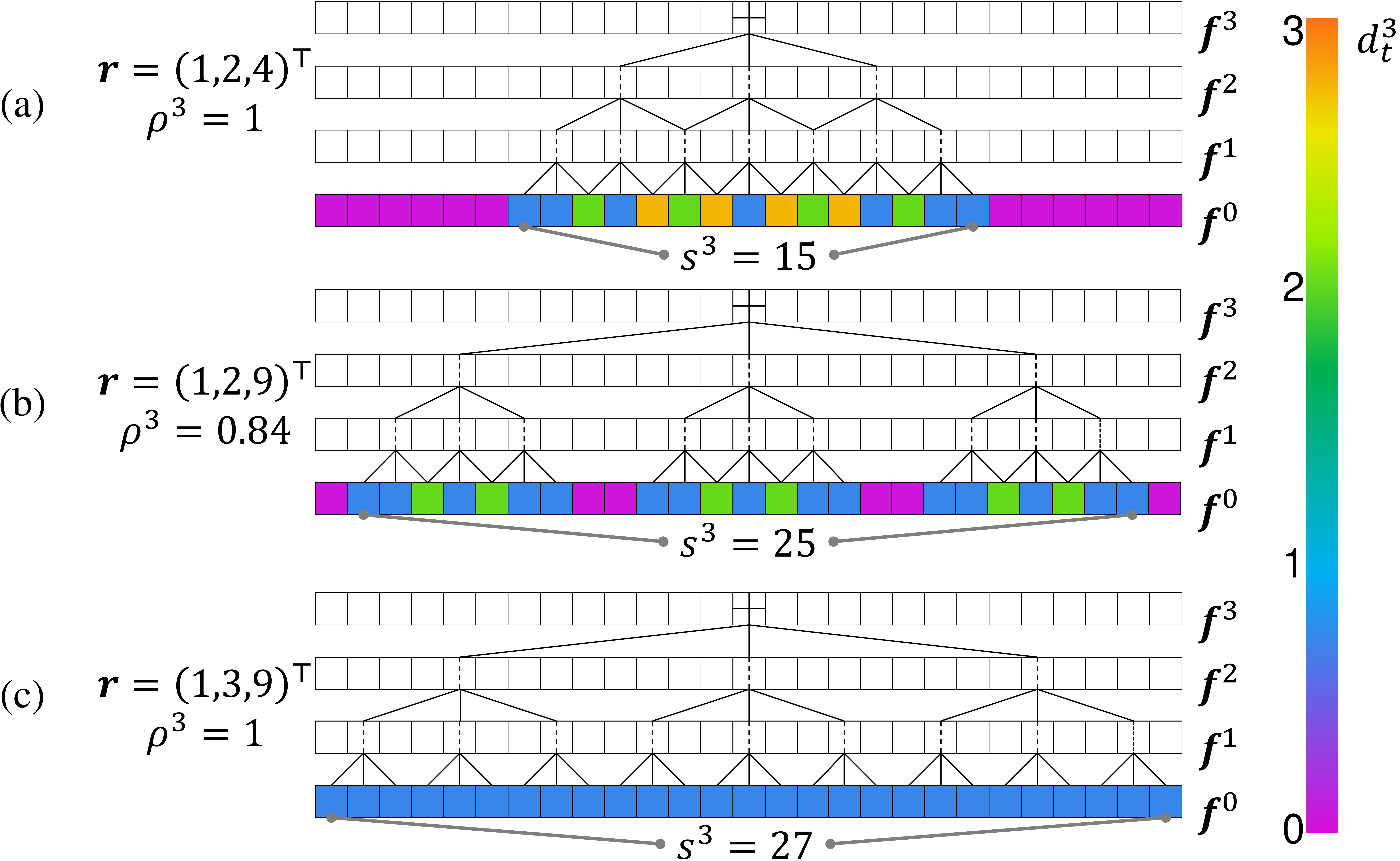}}
    \caption{Three 1D receptive field examples with different atrous rate settings for a three-layer network: (a) an atrous rate setting of (1, 2, 4), (b) an atrous rate setting of (1, 2, 9), (c) an atrous rate setting of (1, 3, 9). The colour represents the link number from the bottom/input node to the top central/output node. $\rho ^3$ is the coverage ratio defined by (\ref{eq-linkratio}), $\textit{\textbf{r}}$ is the atrous rate array, $s^3$ is the receptive field size, $\textit{\textbf{f}}^{(1\sim3)}$ is the 1D feature map, $\textit{\textbf{f}}^{0}$ is the 1D input image, $\textit{d}^3_t$ is the receptive field of $\textit{{f}}_0^{3}$, these notations are explained and used in Sec. \ref{sec: Method-rate}.}
    \label{fig: Rate}
\end{figure}

In this section, we focus on optimizing the atrous rate setting which could achieve the largest and fully-covered receptive field with a minimum number of atrous convolutional layers. Before presenting the detailed mathematical derivation, three 1D receptive field examples with three different atrous rate settings are intuitively shown in Fig. \ref{fig: Rate}. In this three-layer network, with an atrous rate setting of (1, 2, 4), a receptive field of 15 is achieved, while with an atrous rate setting of (1, 2, 9), a receptive field of 25 is achieved with a coverage ratio (the ratio of linked input nodes over all input nodes in the receptive field) of 0.84. With the proposed atrous rate setting of (1, 3, 9), the largest receptive field of 27 is achieved with a full coverage - coverage ratio is 1.0. Detailed mathematical proofs are presented below. For simplification, batch size is fixed at 1 here.

Denote $\textit{\textbf{f}}$ and $\textbf{\textit{F}}$ as 1D and 2D image/feature map. With an input feature map. $\textit{\textbf{F}}^{n-1}$ of size ${\rm H}\times{\rm W} \times{\rm c}_{n-1}$, an output feature map $\textit{\textbf{F}}^{n}$ of size ${\rm H}\times{\rm W} \times {\rm c}_{n}$ is calculated by the $n^{\rm th}$ atrous convolutional layer with an atrous rate ${r}_{n}$, where $\textit{\textbf{F}}^{0}\in\mathbb{R}^{{\rm H}\times{\rm W} \times {\rm c}_{0}}$, $\textit{\textbf{F}}^{n}\in\mathbb{R}^{{\rm H}\times{\rm W} \times {\rm c}_{n}}$, $n\in[1,{\rm N}]\cap\mathbb{N}$, and $\textit{\textbf{r}}=({r}_1\cdots{r}_{\rm N})^\top\in\mathbb{N}^{\rm N}$, where ${\rm N}\in\mathbb{Z}_+$ is the total number of atrous convolutional layers. Here ${\rm H}\in\mathbb{N}$ is the feature height and ${\rm W}\in\mathbb{N}$ is the feature width, though these two values are usually equal for medical images. The channel number of feature maps is denoted as ${\textbf{c}}=({\rm c}_0\cdots{\rm c}_{\rm N})^\top\in\mathbb{N}^{\rm N+1}$, and $\textit{\textbf{F}}^{0}$ is the input image. By ignoring the non-linear modules, i.e., relu, and the biases, an equivalent 2D atrous convolution could achieve a backward propagation from $\textit{\textbf{F}}^{n}$ to $\textit{\textbf{F}}^{n-1}$, which can be decomposed into two 1D atrous convolutions \cite{luo2016understanding}, with kernel $\textit{\textbf{v}}^n$ indexed by $t\in\mathbb{Z}$:

\begin{equation}
\label{eq-kernel}
\textit{{v}}_{t}^{n}({k},{{r}}_{n})=\sum_{u=-{({k}-1)}/{2}}^{{({k}-1)}/{2}} {{{w}}_u^n \cdot{\textbf{1}}(\textit{t}-u{{r}}_{n})}
\end{equation}

Here, $k$ is an odd number which represents the kernel size, i.e., 3, 5, or 7. $t$ is the pixel index. ${w}_u^n$, each element of weight matrix ${w}^n\in\mathbb{R}^k$, is a trainable variable. $\textbf{1}(t):\mathbb{Z}\to\{0,1\}$ is an indicator function defined as:
\begin{equation}
\label{eq-delta}
\textbf{1}(t):=\left\{
\begin{array}{rcl}
1 & t=0\\
0 & t\neq0
\end{array}
\right.
\end{equation}
Denote vectors $\textit{\textbf{f}}^0$, $\textit{\textbf{f}} ^n$ as the 1D input image and the $n^{\rm th}$ 1D feature map, both indexed by $t$. $\textit{\textbf{f}}^{0}$ can be calculated from $\textit{\textbf{f}}^{n}$ by:
\begin{equation}
\label{eq-pathdefinition1}
\textit{\textbf{f}}^{0}=\textbf{\textit{v}}^{1}*\cdots*\textbf{\textit{v}}^{n}*\textit{\textbf{f}}^{n}
\end{equation}
Define $\textbf{\textit{d}}^n(k,\textit{\textbf{r}}):=\textit{\textbf{f}}^{0}(\textit{\textbf{f}}^{n}=\textbf{1}(t))$, in which $\textit{\textbf{f}}^{n}=\textbf{1}(t)$ indicates that only the central pixel of $\textit{\textbf{f}}^{n}$ is with a non-zero value (=1). It is calculated as:
\begin{equation}
\label{eq-pathdefinition}
\textbf{\textit{d}}^n(k,\textit{\textbf{r}}):=\textbf{\textit{v}}^{1}*\cdots*\textbf{\textit{v}}^{n}*{\textbf{1}}(t)
\end{equation}
By setting ${w}^n=(1)_{k},\forall n$, vectors consisting of 1, then $\textit{d}_t^n\in\mathbb{N}$, the element indexed by $t\in\mathbb{Z}$, is the link number from $\textit{{f}}_0^{n}$ to the input image's pixel or node. Thus, $\textbf{\textit{d}}^n$ represents the receptive field of $\textit{{f}}_0^{n}$, where its receptive field coverage could be represented by the non-zero element number in vector $\textit{\textbf{d}}^n$:
\begin{equation}
\label{eq-norm0}
    \left\|\textit{\textbf{d}}^n\right\|_0:=\sum_t{\big(1-\textbf{1}(\textit{d}_t^n)\big)}
\end{equation}
and its receptive field size $\textit{s}^n\in\mathbb{N}$ is calculated as:
\begin{equation}
\label{eq-rfsize}
\textit{s}^{n}({k},\textbf{\textit{r}})=1+({k}-1)\sum_{m=1}^{n}{{r}}_m
\quad
\end{equation}
The receptive field coverage ratio of $\textit{{f}}_0^{n}$, denoted by $\rho ^n\in\mathbb{R}_+$, is then defined as:
\begin{equation}
\label{eq-linkratio}
\rho ^n(k,\textit{\textbf{r}}):=\frac{\left\|\textit{\textbf{d}}^n\right\|_0}{\textit{s}^{n}}
\quad
\end{equation}

In order to ensure a fully-covered receptive field, our target is to maximize the receptive field size with a constraint of receptive field coverage ratio:
\begin{equation}
\label{eq-argmaxorigin}
\left.\mathop{\max}_{\textit{\textbf{r}}\in\mathbb{N}^{\rm N}}{\left\{\textit{s}^{\rm N}:{\rho ^{\rm N}=1}\right\}}\right.
\end{equation}
By substituting (\ref{eq-rfsize}) and (\ref{eq-linkratio}) into (\ref{eq-argmaxorigin}), the optimization problem can be converted as:
\begin{equation}
\label{eq-argmaxconvert}
\left.\mathop{\max}_{\textit{\textbf{r}}\in\mathbb{N}^{\rm N}}{\left\{{\left\|\textit{\textbf{d}}^{\rm N}\right\|_0}:{\left\|\textit{\textbf{d}}^{\rm N}\right\|_0=1+({k}-1)\sum_{n=1}^{\rm N}{r}_n}\right\}}\right.
\end{equation}

The total link number from $\textit{{f}}_0^{n}$ to $\textit{\textbf{f}}^{0}$ is represented by:
\begin{equation}
\label{eq-pathnumber}
\left\|\textit{\textbf{d}}^n\right\|_1 = \sum_t{\textit{{d}}_t^n} = (k)^{n}
\end{equation}
where $(k)^{\rm n}$ represents an exponent calculation. It is the upper bound of $\left\|\textit{\textbf{d}}\right\|_0$ because: 
\begin{equation}
\label{eq-inequation}
\left\|\textit{\textbf{d}}\right\|_0\leq\left\|\textit{\textbf{d}}\right\|_1,\forall{d}_t\in\mathbb{N},\forall{t}\in\mathbb{Z}
\end{equation}
where 
\begin{equation}
\label{eq-equation}
\left\|\textit{\textbf{d}}\right\|_0=\left\|\textit{\textbf{d}}\right\|_1 \Leftrightarrow {{d}}_t\in\{0,1\}, \forall t\in\mathbb{Z}
\end{equation}
We assume that the (\ref{eq-equation}) holds. By substituting this into the constraint of (\ref{eq-argmaxconvert}):
\begin{equation}
1+({k}-1)\sum_{n=1}^{\rm N}{r}_n = (k)^{\rm N}
\end{equation}
This is a sum of geometric progression; one solution can be obtained as:
\begin{equation}
\label{eq-maxatrousrates}
{\textbf{\textit{r}}^\prime}=\begin{pmatrix}1&\cdots&({k})^{n-1}&\cdots&({k})^{\rm N-1}\end{pmatrix}
^\top
\end{equation}
It satisfies a uniformly covered receptive field: 
${{d}}_t^{\rm N}(k,\textit{\textbf{r}}^\prime)=\left\{
\begin{array}{ll}
0 & t\not\in\mathbb{S}\\
1 & t\in\mathbb{S}
\end{array}
\right.
$, where $\mathbb{S}:=[-\frac{s^{\rm N}-1}{2},\frac{s^{\rm N}-1}{2}]\cap\mathbb{Z}$ in 1D and the same in 2D, which satisfies the equivalent condition in (\ref{eq-equation}) and thus is a solution to (\ref{eq-argmaxconvert}).
Therefore, the atrous rate setting of $(k)^{n-1}$ at the $n^{\rm th}$ atrous convolutional layer could lead to the largest and fully-covered receptive field under the condition that the same number of atrous convolutional layers is used. 

\subsection{Truncation effect}
\label{sec: Method-padding}
\begin{figure}
  \centering
  \framebox{\includegraphics[width=0.47\textwidth]{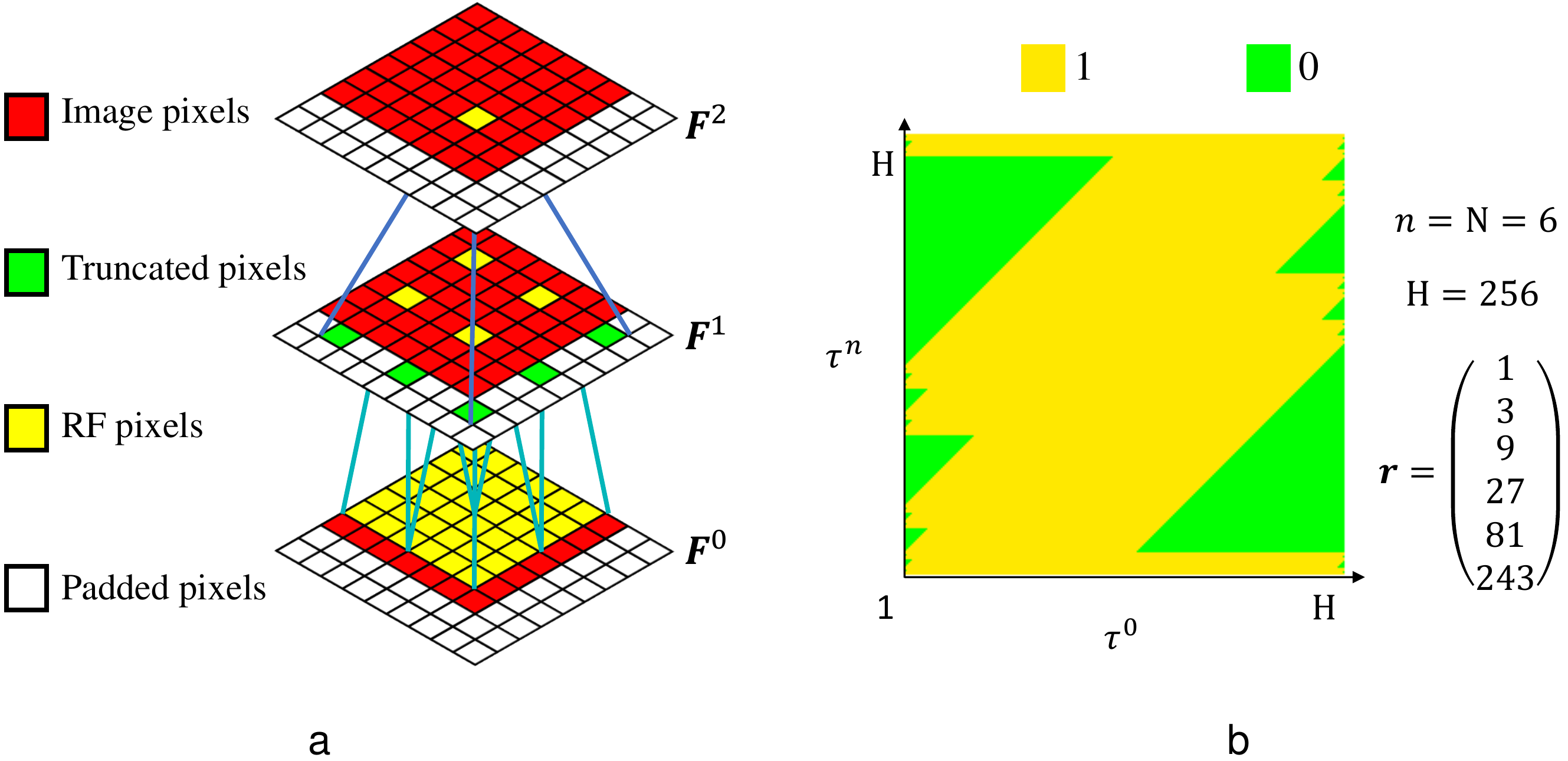}}
  \caption{Illustration of truncation effect: a. back-propagation from $\textit{\textbf{F}}^2$ to $\textit{\textbf{F}}^0$, when $\textit{\textbf{F}}^1$ is calculated from $\textit{\textbf{F}}^2$, only yellow pixels at $\textit{\textbf{F}}^1$ are left, green pixels at $\textit{\textbf{F}}^1$ are truncated, resulting the red pixels at $\textit{\textbf{F}}^0$ are no longer covered by the receptive field of the yellow pixel in $\textit{\textbf{F}}^2$, b. the 1D coverage state of the output node (${\tau}^{n}$) on the input node (${\tau}^{0}$), yellow pixels are covered while green pixels are missed, the total layer number is 6, the 1D image length is 256.}
  \label{fig: Truncation}
\end{figure}

While the mathematical theory can be solved, the practical implementation of this solution must consider that the feature map has edges. Because the calculation results outside the feature map are not stored, some paths are lost, causing the truncation effect. An intuitive illustration is shown in Fig. \ref{fig: Truncation}a. The receptive field of the yellow pixel in $\textit{\textbf{F}}^2$ should be the red and yellow pixels in $\textit{\textbf{F}}^0$. However, the red pixels in $\textit{\textbf{F}}^0$ are not covered due to the truncation of green pixels in $\textit{\textbf{F}}^1$. Mathematical derivations are stated below.

\par
Denote the two boundary pixels as $b$ and $a$, where $b\leq{0}, a\geq{0}$, $a-b={\rm H}-1$. Thus (\ref{eq-pathdefinition}) could be rewritten as (\ref{eq-recursion})
\begin{figure*}
\begin{equation}
\label{eq-recursion}
\begin{matrix}
\textbf{\textit{d}}^n(b,k,\textit{\textbf{r}}):=\\ \quad
\end{matrix}
\begin{matrix}
\underbrace{\Big(\cdots\left[(\textbf{\textit{o}}^{}*\textbf{\textit{v}}^{n})\cdot\varepsilon(t-b)\cdot\varepsilon(a-t)\right]\cdots*\textbf{\textit{v}}^1\Big)\cdot\varepsilon(t-b)\cdot\varepsilon(a-t)}\\n\mbox{-fold atrous convolution with truncation}
\end{matrix}
\end{equation}
\end{figure*}
where $\varepsilon(t):=\left\{
\begin{array}{ll}
0 & t<0\\
1 & t\geq0
\end{array}
\right.
$. Substitute (\ref{eq-kernel}) into (\ref{eq-recursion}), the general term formula for the 1D path number is in (\ref{eq-truncation path}).
\begin{figure*}
\begin{equation}
\label{eq-truncation path}
{{d}}_t^{n}=
\sum_{u_{n}=\max\left(-{(k-1)}/{2},\left\lceil{b}/{{r}_{n}}\right\rceil\right)}^{\min\left({(k-1)}/{2},\left\lfloor{a}/{{r}_{n}}\right\rfloor\right)}
{\left\{\cdots
	\sum_{u_{1}=\max\left(-{(k-1)}/{2},\left\lceil{(b-\sum_{m=2}^{n}{{u_m}{r}_m})}/{{r}_{1}}\right\rceil\right)}^{\min\left({(k-1)}/{2},\left\lfloor{(a-\sum_{m=2}^{n}{{u_m}{r}_m})}/{{r}_{1}}\right\rfloor\right)}{
		\left\{\textbf{1}{\left(t-\sum_{m=1}^{n}{{u_m}{{r}_m}}\right)}
		\right\}
	}\cdots
	\right\}}
\end{equation}

\end{figure*}
We simulate (\ref{eq-truncation path}) in $MATLAB$\textsuperscript{\textcopyright}, an 1D image with length of 256 is put through 6 atrous convolutional layers with an atrous rate of $3^0, 3^1, \cdots, 3^5$ for each, the coverage state of the 256 pixels on the 1D output image is shown in Fig. \ref{fig: Truncation}b. We can see that many pixels are missed (shown in green).

\paragraph{Link with previous work} traditional DCNNs composed of convolutional layers and down-sampling layers are with Gaussian covered receptive field. The path number for nodes at $\textit{\textbf{F}}^{0}$ contribute to $\textit{F}_{0,0}^{\rm N}$ shrinks quickly from the central area to the outer area, which is called Gaussian damage \cite{luo2016understanding}. The weights for the outer area nodes grow during the training, indicating that outer area nodes are also important. A weight initialization with higher weights at the outer area and lower weights at the central area was tried to compensate this Gaussian damage, however, the improvement is limited and unstable \cite{luo2016understanding}. We propose uniformly covered receptive field which could be a solution for Gaussian damage, but with a different purpose - high-resolution feature map propagation.

\subsection{Atrous Convolutional Neural Network}
\label{sec: Method-ACNN}

With the proof in Sec. \ref{sec: Method-rate}, a receptive field of $(k)^{\rm N}$ could be achieved by a block of N atrous convolutional layers. Each node in the receptive field is linked evenly. In this paper, the kernel size of atrous convolutional layers is 3, following the settings used in \cite{simonyan2014very}. A block of N atrous convolutional layers has a receptive field of $(3)^{\rm N}$. We call this block as atrous block and the one specific with N atrous convolutional layers as N-block, here N is expressed in the roman numeral.

\begin{figure*}[thpb]
    \centering
    \framebox{\includegraphics[width=0.7\textwidth]{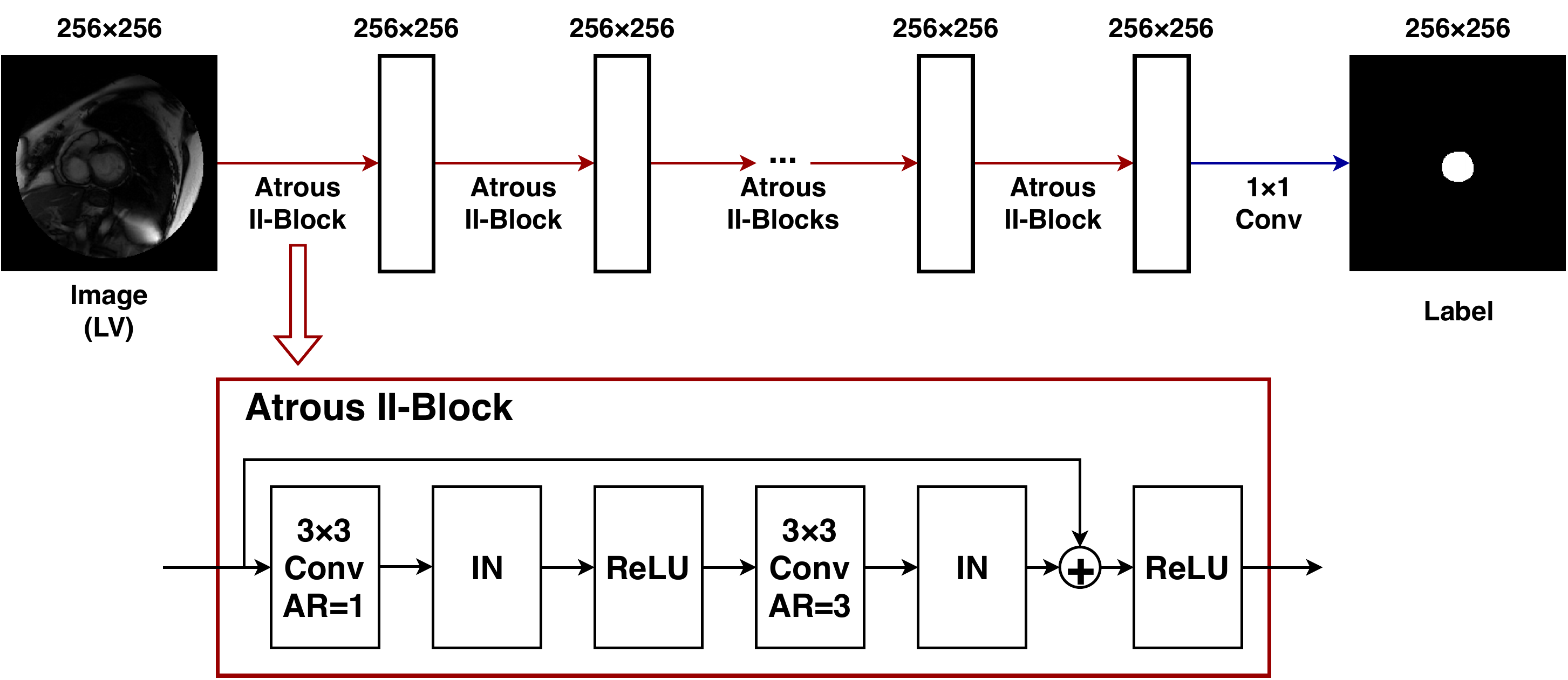}}
  \caption{The network architecture of the proposed ACNN. The number of residual II-blocks is determined by $\frac{(\rm RF-1)}{8}$, $\rm RF$ is the targeted receptive field. AR - atrous rate, $3\times 3$ Conv - atrous convolution with kernel size of 3, $1\times 1$ Conv1 - atrous convolution with kernel size of 1.}
  \label{fig: ACNN}
\end{figure*}

Although a large $\rm N$ indicates larger receptive field, it results in severe truncation effect and less trainable parameters as well. As a trade off, the proposed ACNN is designed into multiple cascaded atrous II-blocks to increase the receptive field linearly by $(3)^{\rm 2}$. For achieving a receptive field of $\rm RF$, $\frac{\rm RF-1}{(3)^{\rm 2}}$ blocks are needed. For solving the gradient vanishing/exploding problems and facilitating back propagation, residual learning \cite{he2016deep} is added while IN is used as the normalization method, following the review in \cite{zhou2019normalization} where IN showed better performance than other normalization method. The final proposed ACNN architecture is shown in Fig. \ref{fig: ACNN}. The number of residual II-blocks - $\frac{\rm (RF-1)}{8}$ is determined by the targeted receptive field.

\subsection{Experimental Setup and Validation}
\label{sec: Data}
Three cardiovascular MRI and CT datasets for RV, LV and aorta segmentation were used for validation.

\paragraph{Right Ventricle (RV)} 37 patients, with different levels of Hypertrophic Cardiomyopathy (HCM) were scanned with a 1.5T MRI scanner (Sonata, Siemens, Erlangen, Germany) \cite{zhou2018real}, involving 6082 images with $10$mm slice gap, $1.5\sim2$mm pixel spacing, $19\sim25$ times frames, and $256\times 256$ image size. Analyze (AnalyzeDirect, Inc, Overland Park, KS, USA) was used to label the ground truth. Rotation from $-30^\circ$ to $30^\circ$ with $5^\circ$ as the interval was used to augment the data. Two groups with 18 and 19 patients were split randomly for cross validations. 

\paragraph{Left Ventricle (LV)} 45 patients, from the SunnyBrook MRI data set \cite{radau2009evaluation} were used, it has 805 images with $256\times 256$ image size. Rotation from $-60^\circ$ to $60^\circ$ with $4^\circ$ as the interval was used to augment the data. Two groups, with 22 and 23 patients respectively, were split randomly for cross validations.

\paragraph{Aorta} 20 patients, from the VISCERAL data set \cite{jimenez2016cloud}, were used, 4631 CT images with $512\times 512$ image size. Rotation from $-40^\circ$ to $40^\circ$ with $10^\circ$ as the interval was used to augment the data. Two groups with 10 patients for each were split randomly for cross validations.

Image intensities were normalized to $0.0\sim 1.0$. Evaluation images were not split. For cross validations, one group was used in the training stage while the other group was used in the testing stage. The kernel size of the last atrous convolutional layer is 1 while the kernel size of all the other atrous convolutional layers is 3. The momentum was set as 0.9. Multiple epoch settings, i.e., 1, 2, or 3 and multiple learning rate schedules, i.e., dividing the learning rate by 5 or 10 at the second or third epoch, indicating an optimal learning schedule that: one epoch was trained and the learning rate was divided by 5 and 25 at 2000 and 4000 iterations respectively. Four initial learning rates: 0.1, 0.05, 0.01, 0.005 were trained for each experiment and the highest accuracy was recorded as the final accuracy to avoid non-optimal hyper-parameter settings. For all experiments, Stochastic Gradient Descent (SGD) was utilized as the optimizer.

Pixel-level softmax was applied to transfer the network outputs into probabilities. Cross-entropy was used as the loss function while IoU was used to evaluate the segmentation accuracy. The worker used was \textit{Titan Xp} (12G memory) with the CPU of Intel® Xeon(R) CPU E5-1650 v4 @ 3.60GHz × 12. The method was implemented with Tensorflow. The process status of the CPU and GPU both influence the training speed. Training all models under exactly the same computer process status is not possible. For a fair speed comparison, the time recorded in this paper is for 100 iterations under the computer process status where all other processes are ended. 

\section{Results}
\label{sec: Result}
The receptive field is important for ACNN performance. In Sec. \ref{sec:result-number}, ACNN with 64, 96, 128 and 192 atrous II-blocks are explored. ACNN-II with a receptive field of double input image size is compared to U-Net \cite{ronneberger2015u} and Deeplabv3+ \cite{chen2018encoder} in Sec. \ref{sec:result-comparison}. Three segmentation and training curve examples are shown in Sec. \ref{sec:result-example} and Sec. \ref{sec:result-loss} respectively.

\subsection{ACNN depth}
\label{sec:result-number}

ACNN with 64, 96, 128 and 192 atrous II-blocks were trained to segment the Fold 2 cross validation of LV. The segmentation mean IoUs, number of parameters and training time for 100 iterations are shown in Tab. \ref{tab:block}. We can see that the segmentation performance increases along the block number of atrous II-blocks, and both the parameter number and training time also increase.

\begin{table}
\label{tab:block}
\caption{The mean IoU, parameter number and training time for 100 iterations of the proposed ACNN with 64, 96, 128 and 192 atrous II-blocks.}
\begin{tabular}{|l|l|l|l|l|}
\hline
\textbf{Data} & \multicolumn{4}{l|}{\textbf{Fold 1 Cross Validation of LV}}\\ \hline
\textbf{Block Number}       & 64      & 96      & 128     & 192     \\ \hline
\textbf{Mean IoU}           & 0.715   & 0.716   & 0.726   & 0.736   \\ \hline
\textbf{Parameter Number}   & 198,928 & 449,456 & 599,984 & 901,040 \\ \hline
\textbf{Training Time}      & 22.1s   & 33.5s   & 42.4s   & 66.8s   \\ \hline
\end{tabular}
\end{table}

\subsection{Comparison to other methods}
\label{sec:result-comparison}

To trade off between the performance and training time, ACNN with a receptive field of double the input image size (512 for the RV and LV while 1024 for the aorta), named ACNN-II, is used to compare with U-Net and Deeplabv3+. The mean IoU, parameter number and model size of the three networks are shown Tab. \ref{tab:comparison}, with validations on the LV, RV and aorta. We can see that the proposed ACNN-II out-performs both the U-Net and Deeplabv3+ in five of the six cross validations. For Fold 2 LV cross validation, ACNN-II achieves very similar mean IoU to the highest one. In addition, the proposed ACNN-II is with much less trainable parameters and model sizes. The performance of Deeplabv3+ is much worse than it was claimed in \cite{chen2018encoder}. This may due to two reasons: 1) it was trained from scratch rather than using pre-trained parameters trained on ImageNet; 2) Deeplabv3+ contains many parameters which causes severe over-fitting in medical image segmentation where the dataset is small. 

\begin{table*}
\label{tab:comparison}
\caption{Comparison of the mean IoU, parameter number and model size of ACNN-II, U-Net and Deeplabv3+ with training on the RV, LV and aorta. Highest values are in bold and blue.}
\centering
\begin{tabular}{|l|l|l|l|l|l|l|l|} 
\hline
\multicolumn{2}{|l|}{\textbf{Data}} & \multicolumn{2}{l|}{\textbf{LV}} & \multicolumn{2}{l|}{\textbf{RV}} & \multicolumn{2}{l|}{\textbf{Aorta}}    \\ 
\hline
\multicolumn{2}{|l|}{\textbf{Cross Validation}} & \textbf{Fold 1} & \textbf{Fold 2} & \textbf{Fold 1} & \textbf{Fold 2} & \textbf{Fold 1} & \textbf{Fold 2}                     \\ 
\hline
\multirow{3}{*}{\textbf{U-Net}} & \textbf{Mean IoU}        & 0.674 & \textcolor{blue}{\textbf{0.722}}           & 0.691 & 0.733           & 0.628 & 0.661                 \\ 
\cline{2-8}
                            & \textbf{Parameter Number} & \multicolumn{6}{l|}{7,782,336}                                                    \\ 
\cline{2-8}
                            & \textbf{Model Size}      & \multicolumn{6}{l|}{62.3 MB}                                                      \\ 
\hline
\multirow{3}{*}{\textbf{Deeplabv3+}} & \textbf{Mean IoU} & 0.523 & 0.509           & 0.429 & 0.489           & 0.513 & 0.639                 \\ 
\cline{2-8}
                            & \textbf{Parameter Number} & \multicolumn{6}{l|}{36,741,392}                                                   \\ 
\cline{2-8}
                            & \textbf{Model Size}      & \multicolumn{6}{l|}{293.9 MB}                                                     \\ 
\hline
\multirow{3}{*}{\textbf{ACNN-II}}   & \textbf{Mean IoU} & \textcolor{blue}{\textbf{0.738}} & 0.715 & \textcolor{blue}{\textbf{0.697}} & \textcolor{blue}{\textbf{0.735}} & \textcolor{blue}{\textbf{0.655}} & \textcolor{blue}{\textbf{0.684}}     \\ 
\cline{2-8}
                            & \textbf{Parameter Number} & \multicolumn{4}{l|}{198,928}                      & \multicolumn{2}{l|}{599,984}  \\ 
\cline{2-8}
                            & \textbf{Model Size}      & \multicolumn{4}{l|}{2.4 MB}                       & \multicolumn{2}{l|}{4.8 MB}   \\
\hline
\end{tabular}
\end{table*}

\begin{figure*}[thpb]
    \centering
    \framebox{\includegraphics[width=\textwidth]{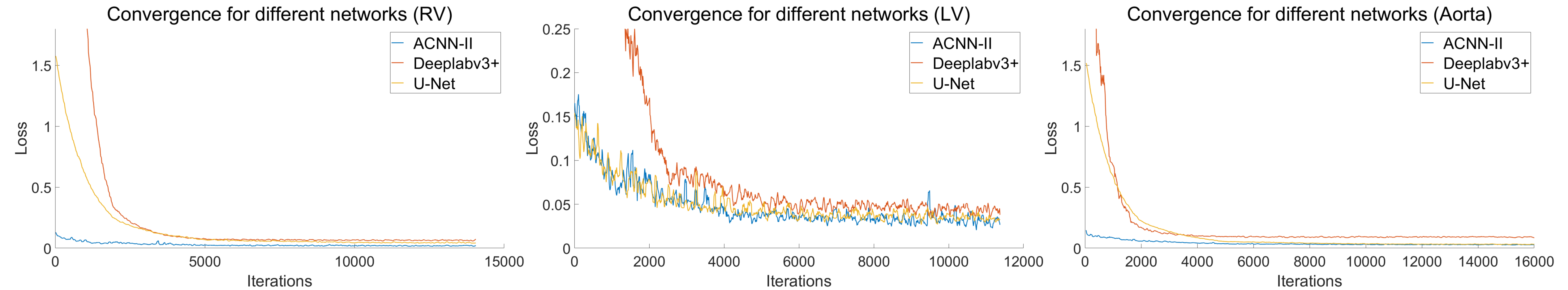}}
  \caption{The loss curves of training ACNN-II, U-Net and Deeplabv3+ for the RV, LV and aortic segmentation.}
  \label{fig:result-loss}
\end{figure*}

\subsection{Segmentation examples}
\label{sec:result-example}

\begin{figure}[thpb]
    \centering
    \framebox{\includegraphics[width=0.47\textwidth]{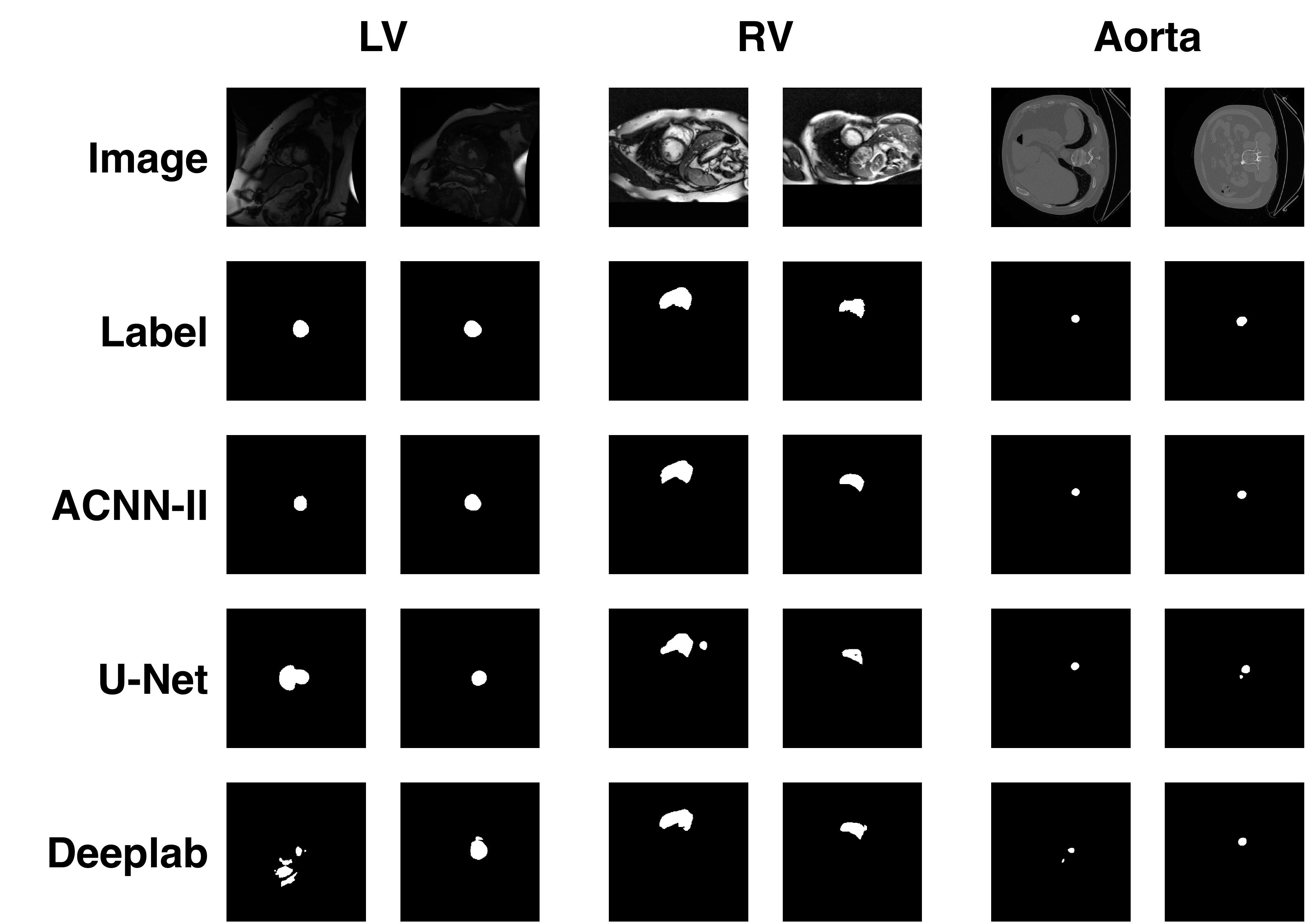}}
  \caption{Two examples of the image, ground truth, and segmentation results of ACNN-II, U-Net and Deeplabv3+.}
  \label{fig:result-example}
\end{figure}

Two segmentation examples are selected from each dataset to show the segmentation details in Fig. \ref{fig:result-example}. We can see that ACNN-II achieves noticeably better visual segmentation results than U-Net and DeeplabV3+ with less false positives.

\subsection{Loss curves}
\label{sec:result-loss}

One training loss is selected from each dataset to show the loss convergence in Fig. \ref{fig:result-loss}. It can be observed that ACNN convergences better and achieves lower loss than U-Net and Deeplabv3+. For the LV and aortic data, the convergence speed is also faster.

\section{Discussion}
\label{sec: Discussion}

An atrous rate setting of $(k)^{n-1}$ at the $n^{th}$ atrous convolutional layer, where $k$ is the kernel size, is proposed. It can achieve the largest and fully-covered receptive field with a minimum number of atrous convolutional layers. Comparison experiments with traditional atrous rate settings, i.e., (1, 2, 4, 8, ...), (1, 2, 5, 9, ...) are not conducted due to: 1) smaller receptive field resulted by traditional atrous rate settings would not definitely indicate lower segmentation accuracy, as a large receptive field may be redundant when the target is small; 2) in addition to the receptive field, complex factors, i.e., the link number of each input node and the trainable parameter number influence the segmentation accuracy too. These complex reasons behind a good segmentation result make it difficult to judge the atrous rate setting from the segmentation accuracy. Hence, in this paper, only detailed mathematical proof and derivations are given.

The proposed ACNN achieves a higher segmentation accuracy than U-Net and Deeplabv3+, but uses much less trainable parameters. We think this achievement comes from the efficient information contained in full resolution feature maps. This advantage is very useful when applying the trained model to mobile devices, as less memory is required. The training time and GPU memory consumption is large in ACNN due to the full resolution feature map calculation. Target specific segmentation DCNNs are not compared in this paper, i.e., Omega-Net proposed for cardiac MRI segmentation \cite{vigneault2018omega} and Equally-weighted Focal U-Net proposed for marker segmentation \cite{zhou2018towards}, as additional algorithms related to the target character is usually applied in these methods and hence these methods usually are not generalizable to other datasets.

Due to the fact that the training time increases along the block number of ACNN, up to a block number of 192 is tested due to the time and resource limitation in Sec. \ref{sec:result-number} and ACNN-II is tested in other sections. In the future, ACNNs with larger block numbers may be further tested. In this paper, ACNN is trained from scratch on three small medical datasets. In the future, the ability of ACNN on large-scale dataset, i.e. PASCAL may be tested, and also its ability for transfer learning.

For a fair comparison, four initial learning rates are explored for each experiment to avoid setting the learning rate less optimally. This process may indicate an optimized accuracy. However, it would not cause unfairness, as it is the same for all experiments. The shown training time is only for 100 iterations and under a clear computer process status. This time could be longer when the computer and GPU are filled with other processes. It can also be different if the implementation is programmed differently. Training configurations, i.e. the momentum and optimizer are selected based on experience. Different results may exist if different training configurations are utilized.

Based on the author's knowledge, all codes were optimized. Further optimization may exist and may influence the recorded memory usage and training time. The applications of the proposed ACNN are not limited to medical image segmentation, but also could be expanded to other pixel-level tasks, which needs further explorations.

\section{Conclusion}
\label{sec: Conclusion}
A new full resolution DCNN - ACNN is proposed for medical image segmentation with the use of cascaded atrous II-blocks, residual learning and IN. A new atrous rate setting is proposed to achieve the largest and fully-covered receptive field with a minimum number of atrous convolutional layers. With much less trainable parameters than that used in the Deeplabv3+ and U-Net, improved accuracy is achieved by even ACNN-II and further improved accuracy can be achieved by deeper ACNN. The derived atrous rate setting contributes to other researches as well. Codes are available at Xiao-Yun Zhou's github.

\section{Acknowledgement}
Thank Qing-Biao Li for the data collection and processing. We thank the support of NVIDIA Corporation with the donation of the Titan Xp GPU used for this research.

\bibliographystyle{IEEEtran}
\bibliography{ICRA2020}

\end{document}